%% file: coling_latex.tex
\pdfoutput=1

\documentclass[11pt]{article}

\usepackage[preprint]{coling}

\usepackage{times}
\usepackage{latexsym}

\usepackage[T1]{fontenc}

\usepackage[utf8]{inputenc}

\usepackage{microtype}

\usepackage{inconsolata}

\usepackage{graphicx}
\usepackage{amsmath}
\usepackage{amssymb}
\usepackage{booktabs}
\usepackage[T1]{fontenc}

\usepackage{array}

\usepackage{multirow}
\usepackage{xcolor}
\usepackage{geometry} %
\usepackage{amsmath}
\usepackage{amssymb}
\usepackage{hyperref}
\usepackage{booktabs}
\usepackage{wrapfig}
\usepackage{url}
\usepackage{xcolor}
\usepackage{algorithm} 
\usepackage{stfloats}
\usepackage{algpseudocode} 

\newcolumntype{P}[1]{>{\centering\arraybackslash}p{#1}}

\title{MedForge: Building Medical Foundation Models Like Open Source Software Development}

\author{
\textbf{Zheling Tan}$^{1*}$, 
\textbf{Kexin Ding}$^{2*}$, 
\textbf{Jin Gao}$^{1}$, 
\textbf{Mu Zhou}$^{2}$, \\
\textbf{Dimitris Metaxas}$^{2}$, 
\textbf{Shaoting Zhang}$^{1}$,
\textbf{Dequan Wang}$^{1\dagger}$\\
$^1$Shanghai Jiao Tong University \quad
$^2$Rutgers University\\
$*$Equal contribution \quad
$\dagger$Corresponding author \\
}

\begin{document}

\maketitle
\begin{abstract}
Foundational models (FMs) have made significant strides in the healthcare domain.
Yet the data silo challenge and privacy concern remain in healthcare systems, hindering safe medical data sharing and collaborative model development among institutions. The collection and curation of scalable clinical datasets increasingly become the bottleneck for training strong FMs. In this study, we propose \textbf{Med}ical \textbf{Fo}undation Models Me\textbf{rg}ing (\textbf{MedForge}), a cooperative framework enabling a community-driven medical foundation model development, meanwhile preventing the information leakage of raw patient data and mitigating synchronization model development issues across clinical institutions. MedForge offers a bottom-up model construction mechanism by flexibly merging task-specific Low-Rank Adaptation (LoRA) modules, which can adapt to downstream tasks while retaining original model parameters. Through an asynchronous LoRA module integration scheme, the resulting composite model can progressively enhance its comprehensive performance on various clinical tasks. 
MedForge shows strong performance on multiple clinical datasets (e.g., breast cancer, lung cancer, and colon cancer) collected from different institutions. 
Our major findings highlight the value of collaborative foundation models in advancing multi-center clinical collaboration effectively and cohesively.
Our code is publicly available at \url{https://github.com/TanZheling/MedForge}.
\end{abstract}



\input{sections/1-intro}

\input{sections/2-related}

\input{sections/4-method}
\input{sections/5-experiments}

\input{sections/6-ablation}

\input{sections/7-discussion}

\input{sections/8-conclusion}

\bibliography{egbib}

\end{document}

%% file: sections/1-intro.tex
\section{Introduction}
\label{sec:intro}

Foundational models (FMs)~\cite{zhang2024data, zhou2023comprehensive} have shown remarkable progress in the healthcare domain, enabling professional-like assessment of disease diagnosis, treatment decision-making, and monitoring~\cite{zhang2023text, wang2022medclip, lu2023mi-zero}. 
Examples include LLaVA-Med~\cite{li2023llava}, Med-PaLM Multimodal~\cite{tu2024towards}, and Med-Flamingo~\cite{moor2023med}, have demonstrated their capacity on question answering, medical image analysis, and report generation.
These studies follow a predominant top-down model development strategy that requires upstream developers to collect data and train models for downstream tasks. 
Consequently, the developed model capabilities are heavily dependent on the training data, limiting their generalization performance in diverse clinical scenarios. 
For instance, Med-Gemini~\cite{yang2024advancing} reveals promising general capabilities in report generation while it lags behind state-of-the-art (SoTA) models on classification tasks, especially for out-of-domain applications. 
This indicates that while the generalizability of the foundation model is promising, more solutions are expected to meet the various specialized clinical needs.

To address these challenges, multi-center data centralization becomes essential to enhance model capacity and robustness across varied clinical scenarios~\cite{rajpurkar2022ai}. 
Centralizing distributed data can significantly improve model training and inference performance.
However, the process of medical data storage, transfer, and aggregation among centers requires extra efforts to ensure data security and system interoperability~\cite{bradford2020international}.
Moreover, a growing concern for patient privacy makes large-scale multi-center data sharing particularly challenging. 
While efforts like federated learning~\cite{wen2023survey, li2020review} can achieve good model performance on local data, the need for synchronized system coordination presents significant challenges, as clients are unable to update asynchronously. This limitation greatly restricts the practical capability of such approaches.
As a result, without a flexible collaboration, medical community still struggles to fully utilize the isolated data and local computation resources for comprehensive medical AI model development. 
To address this dilemma, open-source platforms encourage public data sharing and knowledge integration~\cite{markiewicz2021openneuro, zenodo}.
However, these platforms focus solely on raw data sharing while seldom providing collaborative model training or cooperation between different institutions.
Recently, collaborative learning has emerged as a viable approach for enhancing multi-model robustness~\cite{boulemtafes2020review}. 
For instance, software-like model development~\cite{raffel2023building} mimics software engineering practices by introducing structured workflows, enabling merging, version control, and continuous model integration.
Under this design, model ability can be strengthened with incremental knowledge updates similar to the version updating in software development. 

Although collaborative learning provides a multi-model collaboration, two key challenges remain in the leakage of raw data during collaboration~\cite{huang2023lorahub} and the synchronization of multiple collaborators~\cite{mcmahan2017communication} in the medical AI community. It is still challenging to integrate decentralized, privacy-sensitive data across institutions, leading to under-utilized insights and fragmented knowledge sharing~\cite{kaissis2020secure, rajpurkar2022ai, abdullah2021ethics}.
 To address these challenges, inspired by the collaborative software development, we propose \textbf{Med}ical \textbf{Fo}undation Models Me\textbf{rg}ing (\textbf{MedForge}), a cooperative workflow enabling continuously community-driven foundation model (FM) development.
MedForge enables a lightweight manner for individual centers to share their knowledge among multiple centers, minimizing the burden of data transmission and integration while enhancing model robustness.
Meanwhile, MedForge facilitates asynchronous and flexible collaboration, allowing individual centers to continuously update and improve medical FMs without the need for real-time synchronization.
Similar to open-source software development, MedForge incrementally updates medical knowledge and follows a sustainable model development scheme. 
This key design emphasizes a bottom-up construction of a multi-task medical FM, allowing downstream users to collaboratively build, refine, and update the upstream model according to their local resources. Our major contributions of MedForge are as below: 
\begin{enumerate}
    \item[$\bullet$] We introduce a collaborative workflow to promote the merging scheme of open-source software development. Our proposed MedForge allows distributed clinical centers to asynchronously contribute to comprehensive medical model construction while reducing transmitting costs among centers and avoiding the leakage of raw data, thus enhancing the utilization of private resources in the healthcare system. 
    \item[$\bullet$] We propose two effective knowledge-merging strategies for the asynchronous branch contribution. The MedForge-Fusion strategy updates the plugin module parameters of the main model during the merging phase, whereas the MedForge-Mixture strategy integrates the output of the plugin module by memorizing each contributor's coefficient. These strategies make MedForge more flexible and versatile. MedForge-Fusion is friendly to implement, while the MedForge-Mixture offers better performance and robustness.
    \item[$\bullet$]  We comprehensively evaluate model merging strategies to accumulate medical knowledge among multiple branch plugin modules. MedForge yields superior performance on medical classification tasks compared to other collaborative baselines across multiple datasets. We demonstrate the robustness of MedForge by shuffling the task order and evaluating various configurations of plugin modules and dataset distillation methods.
\end{enumerate}

%% file: sections/2-related.tex
\section{Related Work}
\label{sec:related}
\subsection{Collaborative Systems}
In the era of rapid growth in medical foundational models~\cite{huang2023visual,wang2022medclip, zhang2024data}, the top-down model development paradigm limits model capabilities by heavily relying on the resources available to the model builders. 
Such paradigm often restricts the potential of these models, as they cannot effectively utilize the diverse, private, and decentralized resources that exist within the broader medical community.
In contrast, collaborative systems present a promising alternative, offering a more flexible approach to model development.

Collaborative systems enable institutions to share knowledge by allowing distributed collaborators to contribute to a common goal~\cite{boulemtafes2020review}. 
To further protect patient privacy, federated learning (FL)~\cite{mcmahan2017communication} was proposed to alleviate such privacy concerns as server aggregating parameter updates from multiple clients without sharing their local data. 
While subsequent optimizations, such as aggregation algorithms~\cite{mcmahan2017communication, zhao2018federated, li2020federated}, secure learning~\cite{hardy2017private, xie2021crfl}, fairness improvements~\cite{sharma2022federated, zhao2022dynamic} and its application in medicine~\cite{kumar2024privacy}, have enhanced the capacity and applicability of FL, its real-world flexibility remains limited. This is primarily due to the need for synchronous updates, which require the server and clients to stay in sync, or model updates will be blocked.
This synchrony issue can be mitigated by open-source software platforms (e.g., GitHub~\cite{github}), allowing independent contributions from individual developers asynchronously. Such an asynchronous scheme enables faster iteration and the integration of specialized expertise, thus offering a more flexible and scalable approach.

Unlike synchronous collaboration, asynchronous collaboration does not require collaborators to work simultaneously and collaborators can individually complete their updates.
While the concept of asynchronous collaboration has been widely used in software development, its machine-learning applications remain limited~\cite{kandpal2023git, raffel2023building}. 
With the rise of global collaboration, large models~\cite{sahajBERT, le2023bloom} are usually co-developed by collaborators given various levels of data availability. However, this collaborative scheme requires the aggregation of local data and online synchronous cooperation of developers.
Software-like model update system~\cite{raffel2023building} alleviates the synchronous problem, where models are updated incrementally, similar to software development, by introducing merging and version control to model development.
However, the existing collaborative version control system~\cite{kandpal2023git} fails to address the complexities of medical scenarios because of the heavy dependency on plain parameter averaging across the full model without accounting for the varying requirements of different tasks.
To respond, we propose MedForge, which enables an asynchronous collaborative system and ensures strong robustness toward a continuous, community-driven enhancement of medical models while overcoming potential data leakage.

\input{assets/img/overview}

\subsection{Model Merging}
In collaborative systems, proper model merging becomes increasingly vital for improving model knowledge integration from multiple sources in a resource-limited environment~\cite{li2023deep, yang2024model, goddard2024arcee}. Conceptually, model merging strategies can be categorized into entire model merging and partial model merging.

Entire model merging involves combining multiple model parameters to participate in the merging process by several means. Entire model merging can be viewed as an optimization problem~\cite{Matena_Raffel_2021, jin2022dataless, mavromatis2024packllm} or an alignment problem~\cite{ainsworth2022git, jordan2022repair, xu2024training, ainsworth2022git}, each offering unique advantages depending on the task at hand.
In the optimization-based approach, the goal is to find the best combination of multiple models to enhance performance and efficiency. For instance, using Fisher information approximation~\cite{Matena_Raffel_2021}, the optimization-based model merging can be interpreted as selecting parameters that maximize the joint likelihood of the models' posterior distributions. The optimization of model merging can also be guided by minimizing the prediction differences between the merged model and individual models~\cite{jin2022dataless}. 
With the development of large language models (LLM), optimization-based method is used to fuse multiple LLMs at test-time by minimizing perplexity over the input prompt~\cite{mavromatis2024packllm}.
To highlight, optimization-based methods are beneficial for scenarios requiring enhanced model performance and efficiency to integrate model parameters, while alignment-based methods~\cite{ainsworth2022git, jordan2022repair} are better suited for maintaining consistency and interpretability, facilitating critical information sharing across models.
For example, a training-free model merging strategy aligns relevant models by using a similarity matrix of their representations in both activation and weight spaces~\cite{xu2024training}.
Further, the alignment between the independently trained model and a reference model not only works for models with the same architecture but also for arbitrary model architectures~\cite{ainsworth2022git}.
In summary, the entire model merging methods can effectively integrate existing models into a merged model with enhanced functionality. However, they could lead to increased computational complexity and reduced flexibility, making them less scalable and harder to implement across diverse tasks.

Partial model merging refers to combining only specific components or layers of models to improve model merging efficiency and decrease the computational cost. 
Such specific components can come from the same network~\cite{kingetsu2021neural}, where the original network is divided into subnetworks for different purposes, and these subnetworks can then be recombined for new tasks.
Additionally, modules can originate from different functional networks and be merged using various strategies. For instance, arithmetic operations are applied in \cite{zhang2023composing} to fuse parameter-efficient modules.
While merging modules from different networks provides flexibility, the process also requires a selection strategy to ensure the resulting model aligns with the specific needs of the inference stage. 
The selection strategies are commonly designed based on the similarity of task~\cite{lv2023parameter} and domain clustering performance~\cite{chronopoulou2023adaptersoup}. Alternatively, the mixture-of-experts methods use a routing strategy to select appropriate component modules~\cite{ponti2023combining}. However, these strategies often require significant time and computational resources to filter through a large model pool. 
In contrast, LoRAHub~\cite{huang2023lorahub} offers a more lightweight approach, combining various LoRA modules for different tasks with minimal model training. Nevertheless, LoRAHub lacks flexibility for incremental updates, especially when handling unseen tasks.

Although the existing model merging approaches effectively combine the capabilities of individual models, these approaches often rely on raw data, leading to potential privacy risks. Our proposed MedForge emphasizes the prevention of raw data usage, which is particularly crucial in medical scenarios. Additionally, MedForge offers an extensible capability for incremental learning, enabling continuous model improvement.

%% file: assets/img/overview.tex
\begin{figure*}[ht]
    \centering
    \includegraphics[width=\textwidth]{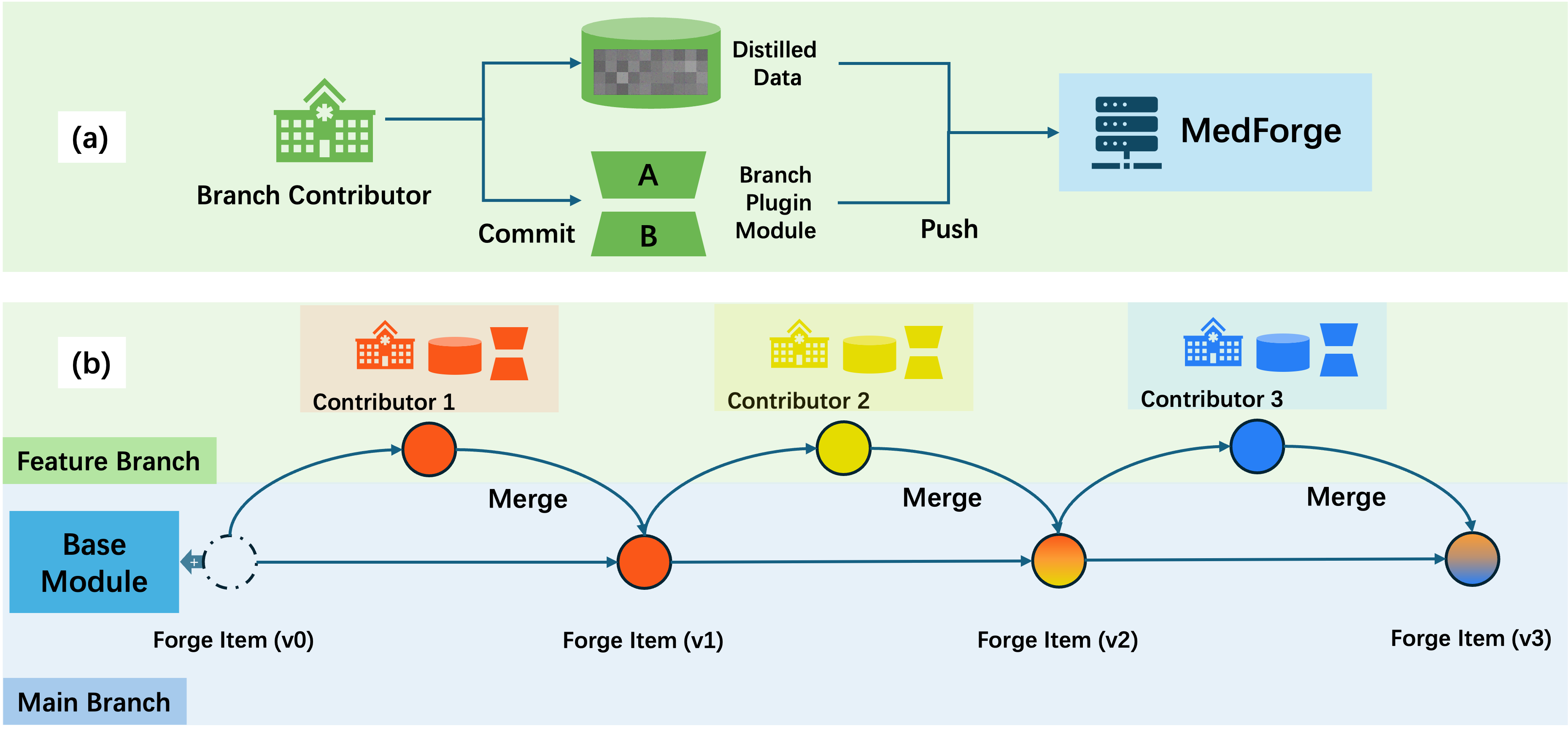}
    \caption{\textbf{The overview of MedForge.} \textbf{(a) Feature branch development.} Branch contributor should commit its branch plugin module and distilled data, then push them to MedForge. MedForge will merge the feature branch with the main branch. In our experiments, we adopt LoRA module as the plugin module. \textbf{(b) Merging stage.} Branch contributors can asynchronously commit and push their branch plugin modules and the distilled datasets to the main branch. Forge items of the main branch will be updated to equip the main branch model with new capabilities.}

    \label{fig:overview}
\end{figure*}

%% file: sections/4-method.tex
\section{Method}

In Fig. \ref{fig:overview}, we illustrate two major stages of MedForge for collaborative model development, including feature branch development (Sec~\ref{branch}) and model merging (Sec~\ref{forging}). In the feature branch development, individual contributors (i.e., medical centers) could make individual knowledge contributions asynchronously. Our MedForge allows each contributor to develop their own plugin module and distilled data locally without the need to share any private data. In the model merging stage, MedForge enables multi-task knowledge integration by merging the well-prepared plugin module asynchronously. This key integration process is guided by the distilled dataset produced by individual branch contributors, resulting in a generalizable model that performs strongly among multiple tasks.

\subsection{Preliminary}
\label{pre}
In MedForge, the development of a multi-capability model relies on the multi-center and multi-task knowledge introduced by branch plugin modules and the distilled datasets.
The relationship between the main base model and branch plugin modules in our proposed MedForge is conceptually similar to the relationship between the main repository and its branches in collaborative software version control platforms (e.g., GitHub~\cite{github}). 
To facilitate plugin module training on branches and model merging, we use the parameter-efficient finetuning (PEFT) technique~\cite{hu2021lora} for integrating knowledge from individual contributors into the branch plugin modules. 

\subsubsection{Parameter-efficient Finetuning}
Compared to resource-intensive full-parameter finetuning, parameter-efficient finetuning (PEFT) only updates a small fraction of the pretrained model parameters to reduce computational costs and accelerate training on specific tasks. These benefits are particularly crucial in medical scenarios where computational resources are often limited.
As the representative PEFT technique, LoRA (Low-Rank Adaptation)~\cite{hu2021lora} is widely utilized in resource-constrained downstream finetuning scenarios. In our MedForge, each contributor trains a lightweight LoRA on a specific task as the branch plugin module. LoRA decomposes the weight matrices of the target layer into two low-rank matrices to represent the update made to the main model when adapting to downstream tasks. If the target weight matrix is $W_0 \in R^{d \times k}$, during the adaptation, the updated weight matrix can be represented as $W_0+\Delta W=W_0+B A$, where $B \in \mathbb{R}^{d \times r}, A \in \mathbb{R}^{r \times k}$ are the low-rank matrices with rank $r \ll  \min (d, k)$ and $AB$ constitute the LoRA module.

\subsubsection{Dataset Distillation}
Dataset distillation~\cite{wang2018dataset, yu2023dataset, lei2023comprehensive} is particularly valuable for medicine scenarios that have limited storage capabilities, restricted transmitting bandwidth, and high concerns for data privacy~\cite{li2024dataset}. 
We leverage the power of dataset distillation to synthesize a small-scale distilled dataset from the original data.

The distilled datasets serve as the training set in the subsequent merging stage to allow multi-center knowledge integration. Models trained on this distilled dataset maintain comparable performance to those trained on the original dataset (\ref{tab:main_res}). Moreover, the distinctive visual characteristics among images of the raw dataset are blurred (see \ref{fig:overview}(a)), which alleviates the potential patient information leakage. 

To perform dataset distillation, we define the original dataset as $\mathcal{T}=\{x_i,y_i\}^N_{i=1}$ and the model parameters as $\theta$. The dataset distillation aims to synthesize a distilled dataset ${\mathcal{S}=\{{s_i},\tilde{y_i}\}^M_{i=1}}$ with a much smaller scale (${M \ll N}$), while models trained on $\mathcal{S}$ can show similar performance as models trained on $\mathcal{T}$. 
This process is achieved by narrowing the performance gap between the real dataset $\mathcal{T}$ and the synthesized dataset $\mathcal{S}$. In MedForge, we utilize the distribution matching (DM)~\cite{zhao2023dataset}, which increases data distribution similarity between the synthesized distilled data and the real dataset
The distribution similarity between the real and synthesized dataset is evaluated through the empirical estimate of the Maximum Mean Discrepancy (MMD)~\cite{gretton2012kernel}:
\begin{equation}
\mathbb{E}_{\boldsymbol{\vartheta} \sim P_{\vartheta}}\left\|\frac{1}{|\mathcal{T}|} \sum_{i=1}^{|\mathcal{T}|} \psi_{\boldsymbol{\vartheta}}\left(\boldsymbol{x}_i\right)-\frac{1}{|\mathcal{S}|} \sum_{j=1}^{|\mathcal{S}|} \psi_{\boldsymbol{\vartheta}}\left(\boldsymbol{s}_j\right)\right\|^2
\end{equation}

where $P_\vartheta$ is the distribution of network parameters, $\psi_{\boldsymbol{\vartheta}}$ is a feature extractor. Then the distillation loss $\mathcal{L}_{DM}$ is:
\begin{equation}\scalebox{0.9}{$
\mathcal{L}_{\mathrm{DM}}(\mathcal{T},\mathcal{S},\psi_{\boldsymbol{\vartheta}})=\sum_{c=0}^{C-1}\left\|\frac{1}{\left|\mathcal{T}_c\right|} \sum_{\mathbf{x} \in \mathcal{T}_c} \psi(\mathbf{x})-\frac{1}{\left|\mathcal{S}_c\right|} \sum_{\mathbf{s} \in \mathcal{S}_c} \psi(\mathbf{s})\right\|^2$}
\end{equation}

We also applied the Differentiable Siamese Augmentation (DSA) strategy~\cite{zhao2021dataset} in the training process of distilled data to enhance the quality of the distilled data. DSA could ensure the distilled dataset is representative of the original data by exploiting information in real data with various transformations. The distilled images extract invariant and critical features from these augmented real images to ensure the distilled dataset remains representative.
\input{assets/img/model_arch}
\subsection{Feature Branch Development}
\label{branch}
In the feature branch development stage, the branch contributors are responsible for providing the locally trained branch plugin modules and the distilled data to the MedForge platform, as shown in Fig~\ref{fig:overview} (a).
In collaborative software development, contributors work on individual feature branches, push their changes to the main platform, and later merge the changes into the main branch to update the repository with new features. Inspired by such collaborative workflow, branch contributors in MedForge follow similar preparations before the merging stage, enabling the integration of diverse branch knowledge into the main branch while effectively utilizing local resources.

MedForge consistently keeps a base module and a forge item as the main branch. The base module preserves generative knowledge of the foundation model pretrained on natural image datasets (i.e., ImageNet~\cite{deng2009imagenet}), while the forge item contains model merging information that guides the integration of feature branch knowledge (i.e., a merged plugin module or the merging coefficients assigned to plugin modules). 
Similar to individual software developers working in their own branches, each branch contributor (e.g., individual medical centers) trains a task-specific plugin module using their private data to introduce feature branch knowledge into the main branch. These branch plugin modules are then committed and pushed to update the forge items of the main branch in the merging stage, thus enhancing the model's multi-task capabilities.

\input{assets/img/merge}

Regarding model architecture, MedForge contains a base module and a plugin module (Fig ~\ref{fig:model_arch}). The base module is pretrained on general datasets (e.g., ImageNet) and remains the model parameters frozen in all processes and branches (main and feature branches) to avoid catastrophic forgetting of foundational knowledge acquired from pretraining. Meanwhile, the plugin module is adaptable for knowledge integration and can be flexibly added or removed from the base module, allowing updates without affecting the base model. In our study, we use the pretrained CLIP~\cite{radford2021learning} model as the base module. For the language encoder and projection layer of the CLIP model, all the parameters are frozen, which enables us to directly leverage the language capability of the original CLIP model. For the visual encoder, we apply LoRA on weight matrices of query ($W_q$) and value ($W_v$), following the previous study~\cite{hu2021lora}. To better adapt the model to downstream visual tasks, we apply the LoRA technique to both the visual encoder and the visual projection, and these LoRA modules perform as the plugin module. During the training, only the plugin module (LoRA modules) participates in parameter updates, while the base module (the original CLIP model) remains unchanged. 

In addition to the plugin modules, the feature branch contributors also develop a distilled dataset based on their private local data, which encapsulates essential patterns and features, serving as the foundation for training the merging coefficients in the subsequent merging stage~\ref{forging}. Compared to previous model merging approaches that rely on whole datasets or few-shot sampling, distilled data is lightweight and representative, mitigating the privacy risks associated with sharing raw data. 
We illustrate our distillation procedure in Algorithm~\ref{algorithm:alg1}. In each distillation step, the synthesized data $\mathcal{S}$ will be updated by minimizing $\mathcal{L}_{DM}$.
\input{alg1}

\subsection{MedForge Merging Stage}
\label{forging}
Following the feature branch development stage illustrated in Fig~\ref{fig:overview} (a), branch contributors push and merge their branch plugin modules along with the corresponding distilled dataset into the main branch, as shown in Fig~\ref{fig:overview} (b). Our MedForge allows an incremental capability accumulation from branches to construct a comprehensive medical model that can handle multiple tasks.

In the merging stage, the $i^{th}$ branch contributor is assigned a coefficient $w'_i$ for the contribution of merging, while the coefficient for the current main branch is $w_i$. By adaptively adjusting the value of coefficients, the main branch can balance and coordinate updates from different contributors, ultimately enhancing the overall performance of the model across multiple tasks.
The optimization of the coefficients is done by minimizing the cross-entropy loss for classification based on the distilled datasets. We also add $L1$ regularization to the loss to regulate the weights to avoid outlier coefficient values (e.g., extremely large or small coefficient values)~\cite{huang2023lorahub}. During optimization, following~\cite{huang2023lorahub}, we utilize Shiwa algorithm~\cite{liu2020versatile} to enable model merging under gradient-free conditions, with lower computational and time costs. The optimizer selector~\cite{liu2020versatile} automatically chooses the most suitable optimization method for coefficient optimization. 

In the following sections, we introduce the two merging methods used in our MedForge: Fusion and Mixture. In MedForge-Fusion, the parameters of the branch plugin modules are fused into the main branch after each round of the merging stage. For MedForge-Mixture, the outputs of the branch modules are weighted and summed based on their respective coefficients rather than directly applying the weighted sum to the model parameters. This largely preserves the internal parameter structure of each branch module.

\paragraph{MedForge-Fusion}
In MedForge, forge items are utilized to facilitate the integration of branch knowledge into the main branch.
For MedForge-Fusion, the forge item refers to adaptable main plugin modules. When the $i^{th}$ branch contributor pushes its branch plugin module $\theta'_i=A'_iB'_i$ to the main branch, the current main plugin module $\theta_{i-1}=A_{i-1}B_{i-1}$ will be updated to $\theta_{i}=A_{i}B_{i}$. The parameters of the branch and the current main plugin modules are weighted with coefficients and added to fuse a new version. The $A_i$, $B_i$ are the low-rank matrices composing the LoRA module $\theta_i$. The detailed fusion process can be represented as:
\begin{equation}
\theta_{i}=(w_i A_{i-1}+w'_i A'_i)(w_i B_{i-1}+ w'_i B'_i)
\end{equation}
Where $w_i$ is the coefficient assigned to the current main branch, while $w'_i$ is the coefficient assigned to the branch contributor. After this round of merging, the resulting plugin module $\theta_{i}$ is the updated version of main forging item, thus the main model is able to obtain new capacity introduced by the current branch contributor. When new contributors push their plugin modules and distilled datasets, the main branch can be incrementally updated through merging stages, and the optimization of the coefficients is guided by distilled data.
As shown in Fig.~\ref{fig:merge}, though multiple contributors commit their branch plugin modules and distilled datasets at different times, they can flexibly merge their plugin modules with the current main branch. After each merging round, the plugin module of the main branch will be updated, and thus the version iteration has been achieved.
\input{assets/img/mixmerge}

\paragraph{MedForge-Mixture}
To further improve the model merging performance, inspired by~\cite{zhao2024loraretriever}, we also propose medForge-mixture. For MedForge-Mixture, the forge items refer to the optimized coefficients.
As shown in Fig.~\ref{fig:mixmerge}, for MedForge-Mixture, the coefficient of each branch contributor is acquired and optimized based on distilled datasets. Then the outputs of plugin modules will be weighted combined with these coefficients to get the merged output. 

For each merging round, with branch contributor $i$, the branch coefficient is $w'_i$, the main coefficient is $w_i$, the branch plugin module is $\theta'_i=A'_iB'_i$, and the current main plugin module is $\theta_i=A_iB_i$. With the input $x$, the resulted MedForge-Mixture output can be represented as:
\begin{equation}
y_{i}=w_i A_{i-1} B_{i-1} x+w'_i A'_i B'_i x
\end{equation}

In this way, MedForge encourages additional contributors as the workflow supports continuous incremental knowledge updates.

Overall, both MedForge merging strategies greatly improve the communication efficiency among contributors. We use this design to build a multi-task medical foundation model that enhances the full utilization of resources in the medical community. For the MedForge-Fusion strategy, the main plugin module is updated after each merging round, thus avoiding storing the previous plugin modules and saving space. Meanwhile, the MedForge-Mixture strategy avoids directly updating the parameters of each plugin module, thus preserving their original structure and preventing the introduction of additional noise, which enhances the robustness and stability of the models.

%% file: assets/img/model_arch.tex
\begin{figure}[t]
    \centering
    \includegraphics[width=\linewidth]{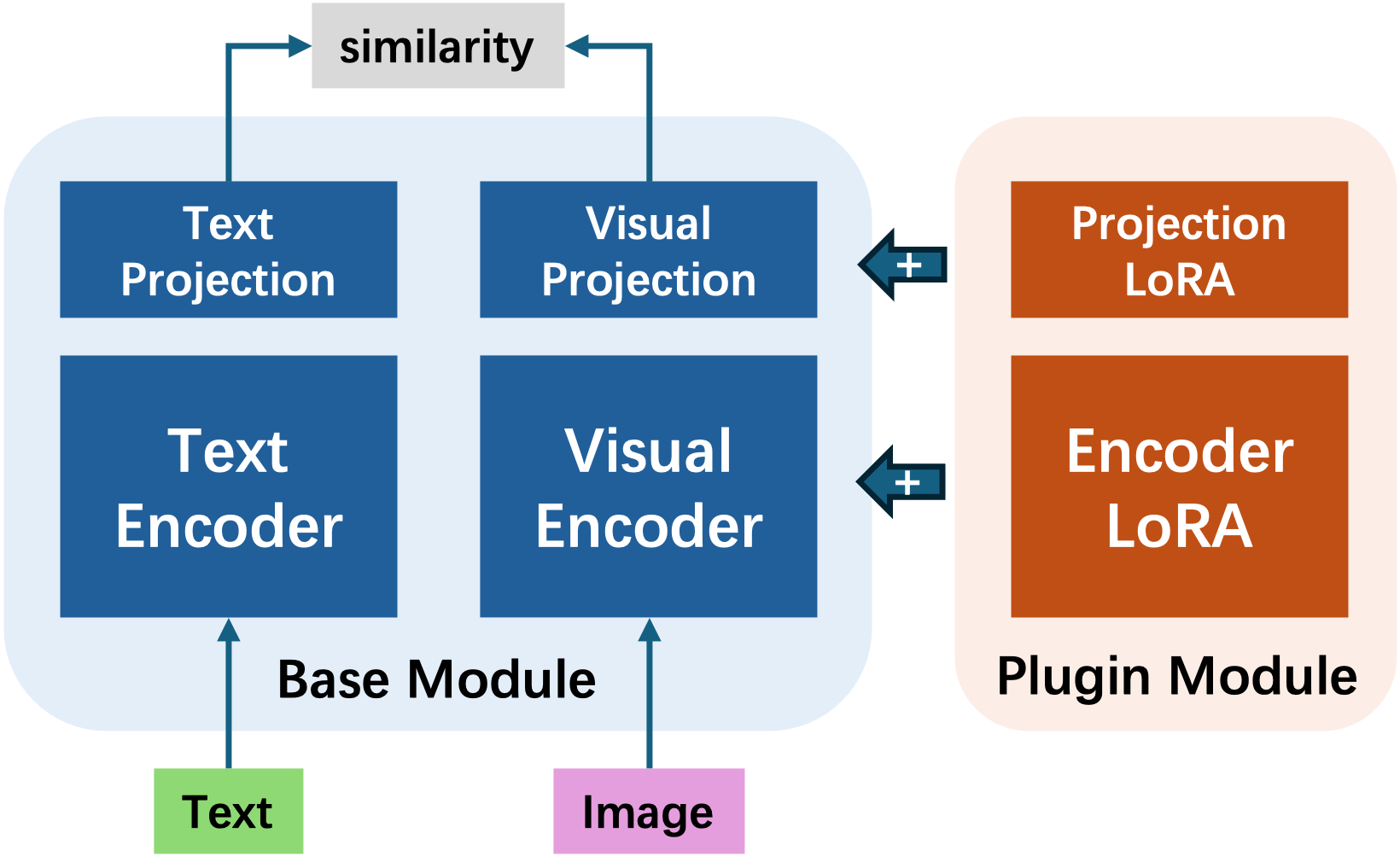}
    \caption{\textbf{Main model architecture.} We adopt CLIP as the base module and attach LoRA modules to the visual encoder and visual projection as the plugin module. During all the procedures, only the plugin modules are tuned while the rest are frozen. We get the classification result by comparing the cosine similarity of the visual and text embeddings.}
    \label{fig:model_arch}
\end{figure}

%% file: assets/img/merge.tex
\begin{figure*}
    \centering
    \includegraphics[width=\textwidth]{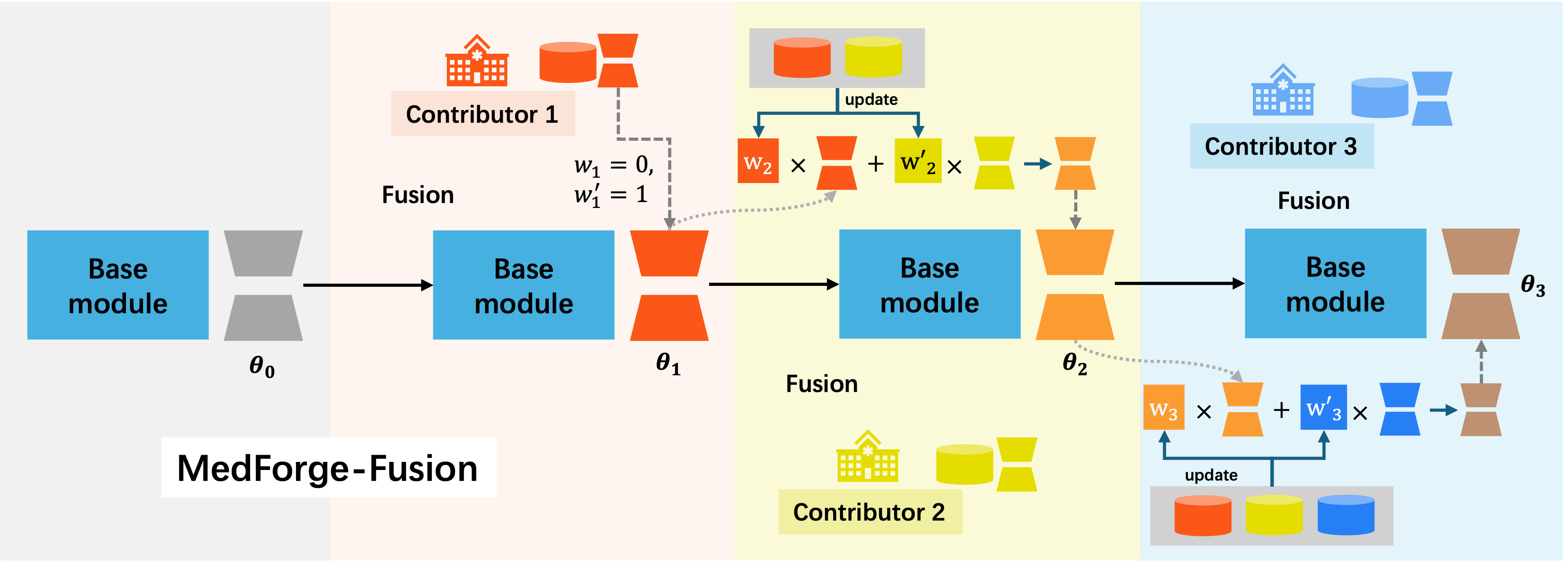}
    \caption{\textbf{The detailed methodology of the proposed Fusion.} Branch contributors can asynchronously commit and push their branch plugin modules and the distilled datasets. the plugin modules will then be weighted fused to the current main plugin module.}

    \label{fig:merge}
\end{figure*}

%% file: alg1.tex
\begin{algorithm} 
    \caption{Dataset Distillation of Branch Private Data}    
    \label{algorithm:alg1}       
    \begin{algorithmic}[1] 
    \Require $\theta_{main}$: pretrained main model parameters; $E$: number of distillation steps; 
    \Ensure distilled data $S$
    \State Initialize ${S=\{{s_i},\tilde{y_i}\}^M_{i=1}}$ randomly, 
    
    \For {distillation steps $e=1$ to $E$}
    \State $\theta \leftarrow \theta_{main} + \theta_{LoRA}$, $\theta_{LoRA}$ is randomly initialized
    \State $\psi_{\boldsymbol{\vartheta}}$ $\leftarrow$ image feature extractor of $\theta $
    \State Sample a minibatch $B^\mathcal{T} \sim \mathcal{T}$ and $B^\mathcal{S} \sim \mathcal{S}$
    \State Compute $\mathcal{L}_{DM}(\mathcal{T},\mathcal{S},\psi_{\boldsymbol{\vartheta}})$
    \State $\mathcal{S} \leftarrow SGD\ (\mathcal{S};\mathcal{L}_{DM})$
    \State Update $\theta_{LoRA}$ with $S$
    \EndFor
    \end{algorithmic} 
\end{algorithm}

%% file: assets/img/mixmerge.tex
\begin{figure*}[t]
    \centering
    \includegraphics[width=\textwidth]{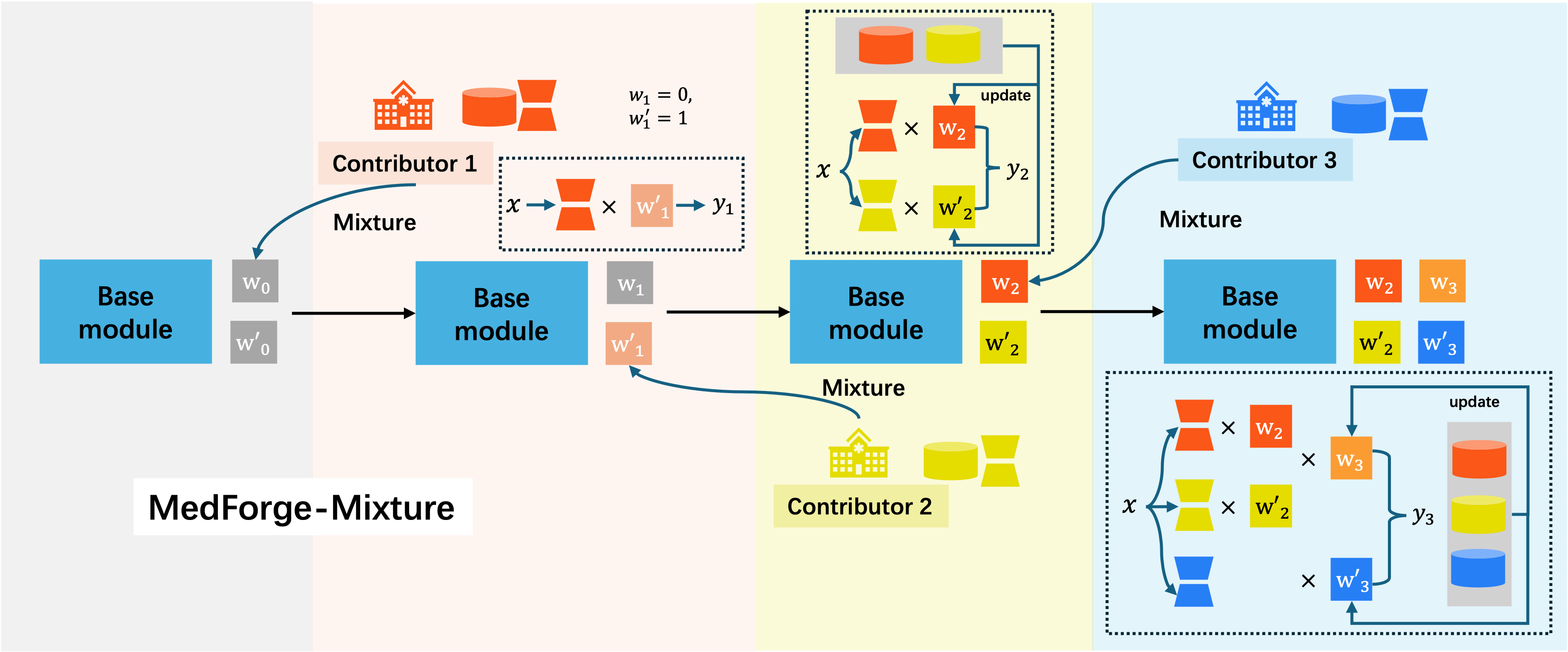}
    \caption{\textbf{The detailed methodology of the proposed Mixture.} Branch contributors can asynchronously commit and push their branch plugin modules and the distilled datasets. the outputs of different plugin modules will be weighted aggregated. The weights of each merging step will be saved.}

    \label{fig:mixmerge}
\end{figure*}

%% file: sections/5-experiments.tex
\section{Experiments}
\subsection{Datasets}
\paragraph{BreakHis}
BreakHis~\cite{spanhol2015dataset} is a breast cancer histopathology dataset with 7,909 histopathological image patches acquired from 82 patients. We utilize images with magnification in 40${\times}$ to achieve a high-resolution image quality, which can reveal the valuable disease patterns among cancer tissues. We randomly selected 70\% of the patients and used their image samples as the training set. The image samples from the remaining patients were used as the test set. All of the images are first center cropped to 460${\times}$460 pixels as the raw images are 700${\times}$460, then resized to 224 ${\times}$ 224. 

\paragraph{LC25000}
LC25000~\cite{borkowski2019lung} contains 25,000 images of lung and colon cancer histopathology with five clinical categories. In our experiment, we include benign lung tissues and two subtypes of lung cancer for classification tasks, including lung adenocarcinomas and lung squamous cell carcinomas. The images are resized to 224 ${\times}$ 224. 70\% of the images are randomly selected as the training data, and the rest of the images serve as an evaluation set. 

\paragraph{MedFMC-Colon}
MedFMC~\cite{wang2023medfmc} is established to evaluate the adaptation performance of foundation models on real-world medical image classification tasks. In our study, we evaluate the model performance on histopathological colon tumor tissue classification task and follow the dataset split setting released by previous study~\cite{wang2023medfmc}. The MedFMC-Colon dataset contains 10,009 images. We first resized the images to 224 ${\times}$ 224 to keep with other datasets, then applied a random horizontal flip to augment the data following the previous study~\cite{wang2023medfmc}.
\input{assets/tab/main_res}
\input{assets/tab/order}
\input{assets/tab/rebuttal}
\input{assets/tab/lora}
\subsection{Implementations}
We define CLIP ViT-B/16 as the base module, with input image size 224 ${\times}$ 224, patch size 16 ${\times}$ 16, and the output dimension of the image and text projection layer are both 512. For the branch plugin module, we implemented LoRA tuning based on the Huggingface PEFT library, the default hyperparameter rank r and lora\_alpha are fixed to 16, lora\_dropout is 0.1, and bias is set to be none.
The classification prediction is determined by calculating the cosine similarity between the visual and text embeddings. Specifically, we first compute the feature embedding of the image and the candidate texts through the image encoder and text encoder. The cosine similarity between these embeddings is then calculated to determine the most probable image-text pairing. The image-text pair with the highest cosine similarity is selected as the predicted classification, indicating the closest match between the visual content and the corresponding textual description.
In our experiments, the labels of the images in the dataset are converted as the text input to the model's text encoder. Specifically, in BreakHis, the labels are 'benign breast tissues' and 'malignant breast tumors'. For LC25000, the subtype labels are 'lung squamous cell carcinomas', 'lung adenocarcinomas', and 'benign lung tissues'. While in MedFMC-Colon, the labels are 'negative colon tumor' and 'positive colon tumor'.
For coefficient optimization of merging contribution, the Cobyla~\cite{powell1994direct} strategy is automatically selected by the Shiwa algorithm~\cite{liu2020versatile} in our experiment setting. The default number of updating iterations is set to 40, and the initial weight is 0.5.

In our study, we use the above public medical datasets to mimic the scenario of the private data held by each medical center. The individual center contributors could only use their private dataset to train a branch plugin module and the corresponding distilled dataset for the specific task. Then, these individual centers contribute the best-performed branch plugin modules and the distilled datasets for the subsequent merging stage, aiming to develop an integrated model without raw private data leakage. All the results shown below are repeated with three random seeds.

\subsection{Baselines}

To demonstrate the performance of our method, we compared two types of baselines: the first is individual baselines, which provide benchmarks for single-task performance without collaboration. The second is collaborative baselines, which focus on evaluating the collaborative performance of merging knowledge from multiple contributors.

\subsubsection{Individual Baselines}
The individual baselines highlight model performance on single tasks without collaborative efforts in MedForge. In individual baselines, each branch contributor operates independently on their own dataset using the same base model and plugin module as our proposed MedForge. We include two types of individual baselines, i.e., single-task LoRA tuning and dataset distillation. These baselines are considered upper bounds because they are trained exclusively on raw or distilled data for a single task without the need for trade-offs for handling multiple tasks, as in the case of MedForge.

\paragraph{Single-task LoRA Tuning}
Single-task LoRA tuning trains LoRA modules specifically based on each raw dataset, making them highly specialized for a single task. As a result, the performance of these specialized models can be considered the upper bound for all the other methods, as they are optimized for one task with a dedicated set of data, achieving the best possible results for that particular task. 
For the training settings, we train 100 epochs optimized by stochastic gradient descent (SGD), with a learning rate of 0.01, a momentum of 0.9, and a weight decay of 0.0005. The batch size is set as 64.

\paragraph{Single-task Dataset distillation}
To determine the capability of collaborative methods and the effectiveness of distilled data, we evaluate the performance of models trained solely on the distilled datasets provided by branch contributors. In our study, distilled data serves as specialized knowledge from private datasets, guiding the knowledge integration in the merging stage. The performance of models trained on distilled data also serves as an upper-bound baseline for comparing the effectiveness of the model merging process.
During the distillation and evaluation, our setting follows the design in the previous study~\cite{cazenavette2023glad}. The default IPC (image per class) of the distilled dataset is set to 20, and the distillation iteration is 5000. In evaluation, we train the randomly initialized branch plugin module based on the distilled images with 1,000 iterations and then test on the original test sets to examine the quality of the distilled data. The best-performed distilled data will be selected by each contributor based on the evaluation of accuracy and AUC on test dataset. We repeat the above setting for each dataset.

\subsubsection{Collaborative Baselines}
We introduce two collaborative baselines that merge knowledge from multiple contributors: LoRAHub~\cite{huang2023lorahub} and ModelSoup~\cite{wortsman2022model}. Compared with the above collaborative-based baseline, our proposed MedForge method highlights the performance advantages of merging knowledge from multiple contributors.

\paragraph{ModelSoup}
We adopt the model parameter averaging strategy proposed by ModelSoup~\cite{wortsman2022model} as one of our model merging baselines.
ModelSoup trained multiple models with various hyperparameters, and then averaged the parameters of various beneficial candidate models. Since in our setting, the plugin modules are trained for different purposes, we fuse the branch plugin modules by averaging their model parameters for ModelSoup baseline.

\paragraph{LoRAHub}
LoRAHub~\cite{huang2023lorahub} serves as an important baseline as it focuses on the merging of multiple task-specific LoRAs. However, it merges synchronously and still relies on raw data. In LoRAHub, before merging, all tasks involved in the merging process require trained LoRA modules and a few-shot sample from the original dataset to train the merging coefficients, resulting in a weighted combination of the LoRA modules.
As LoRAHub was proposed for language tasks, which is different from our settings, we designed two comparative means: one based on the raw training sets (w/o distill) and the other on the distilled datasets (w/ distill). The main difference lies in the data used for training the coefficients. For w/o distill, we use a randomly selected 10\% of the raw dataset to guide the weight optimization, while for w/ distill, we use the distilled data as guidance, similar to MedForge.
In the LoRAHub setting, all contributors can only synchronously work together, which greatly hinders the flexibility of the merging system. In addition, once model merging is completed, the capability to solve the tasks will be fixed. When facing new tasks, retraining the weight assigned for each branch module is required, which significantly increases the operational burden of continuously updating novel knowledge.

\subsection{Evaluation Metrics}
We evaluate the model performance with accuracy (ACC) and area under the receiving operator characteristic curve (AUC). We evaluate the slide-wise prediction results for BreakHis by using the slide ID. We aggregate the patch-wise prediction results by calculating the mean output of the slide patches and applying softmax on them. For LC25000 and MedFMC-Colon, we evaluate their performance with patch-wise ACC and AUC because no corresponding patient or slide information is provided.

\input{assets/img/feature}
\subsection{The Performance of MedForge}
In Table~\ref{tab:main_res}, MedForge has shown superb performance on the integration of the three medical tasks compared to other baselines with asynchronous workflow.

Among multiple datasets, our proposed MedForge-Mixture method outperforms all the other merging baseline models up to about 20\% in ACC and 8\% in AUC (Avg-3 tasks: row 7 (MedForge-Mixture) versus row 3-5 (baselines) ) and even outperform the upper bound of single synthesized dataset training on both ACC (row 7 versus row 2, 0.909 versus 0.864) and AUC (row 7 versus row 2, 0.941 versus 0.91).
Overall, the results demonstrate the strong integration performance of our proposed MedForge workflow with asynchronous merging conditions.
The notable performance of MedForge-Mixture method can be attributed to the use of the distilled dataset as the training basis for optimal coefficients. Additionally, integrating the results of the plugin modules helps maintaining parameter organization within the branch plugin module and avoids noise that could be introduced by the direct parameter operation. MedForge-Mixture also achieves the best performance on both LC25000 and MedFMC-Colon compared to other model merging methods, while comparable good results on BreakHis. These results show the generalizability of MedForge-Mixture among multiple tasks.
Further, the MedForge-Fusion approach shows comparable capabilities to other model merging baselines focusing on knowledge integration or parameter fusion. Fusion-based merging strategy surpasses ModelSoup by 10\% performance on ACC. MedForge-Fusion also has better performance on AUC compared to LoRAHub w/o distill. 
Due to a differential model design, we have observed a difference in performance when using MedForge-Fusion and MedForge-Mixture strategies. MedForge-Fusion fuses the branch plugin module parameters and updates the main plugin modules, while MedForge-Mixture maintains the internal parameters of each plugin module. The former reduces the cost of plugin module memory while the latter introduces less noise to the parameter structure of branch modules.

%% file: assets/tab/main_res.tex
\renewcommand{\arraystretch}{1.2}
\begin{table*}[ht]
\centering
\caption{\textbf{The Evaluation of the Performance of Med-Forge.} Row 1 and Row 2 reflect a supervised-based upper-bound and the evaluation results of the distilled dataset as introduced above.
Row 3-5 show the performance of other model merging methods, including the variants of the LoRAHub technique and the ModelSoup method. Row 6-7 shows the results of our proposed MedForge, including two types of merging approaches. Compared to other fusion methods, MedForge shows superb integrated performance, especially for MixMerge mode.}
\label{tab:main_res}
\resizebox{\textwidth}{!}{
\begin{tabular}{c|c|cc|cc|cc|cc}
\toprule
\multirow{2}{*}{Row}&\multirow{2}{*}{Method}& \multicolumn{2}{c|}{BreakHis} & \multicolumn{2}{c|}{LC25000} & \multicolumn{2}{c|}{MedFMC-Colon} & \multicolumn{2}{c}{Avg-3 tasks} \\
 & & \multicolumn{1}{c|}{ACC} & AUC & \multicolumn{1}{c|}{ACC} & AUC & \multicolumn{1}{c|}{ACC} & AUC & \multicolumn{1}{c|}{ACC} & AUC \\ \hline
1 & LoRA sup & \multicolumn{1}{c|}{$0.914 \pm 0.012$ } & {$0.923 \pm 0.007$ } & \multicolumn{1}{c|}{$0.997 \pm 0.001$} & {$0.996 \pm 0.006$} & \multicolumn{1}{c|}{$0.978 \pm 0.009$} & {$0.997 \pm 0.003$} & \multicolumn{1}{c|}{$0.963 \pm 0.005$} & {$0.972 \pm 0.002$} \\
2 & Distilled data only & \multicolumn{1}{c|}{$0.881 \pm 0.006$} & {$0.883 \pm 0.039$} & \multicolumn{1}{c|}{$0.911 \pm 0.002$} & {$0.974 \pm 0.001$} & \multicolumn{1}{c|}{$0.799 \pm 0.011$} & {$0.872 \pm 0.016$} & \multicolumn{1}{c|}{$0.864 \pm 0.002$} & {$0.910 \pm 0.010$} \\
\hline
3 & LoRAHub w/o distill & \multicolumn{1}{c|}{$0.705 \pm 0.051$} & {$0.641 \pm 0.093$} & \multicolumn{1}{c|}{$\textbf{0.959} \pm 0.02$} &{$ 0.997 \pm 0.003$} & \multicolumn{1}{c|}{$0.825 \pm 0.147$} & {$0.945 \pm 0.034$} & \multicolumn{1}{c|}{$0.830 \pm 0.036$} & {$ 0.861 \pm 0.035$} \\
4 & LoRAHub w/ distill & \multicolumn{1}{c|} {$0.654 \pm 0.039$} & {$0.804 \pm 0.184$} & \multicolumn{1}{c|}{$0.956 \pm 0.019$} &{$ 0.997 \pm 0.001$} & \multicolumn{1}{c|}{$0.819 \pm 0.119$} & {$0.936 \pm 0.041$} & \multicolumn{1}{c|}{$0.810 \pm 0.026$} & {$ 0.912 \pm 0.058$} \\
5 & ModelSoup & \multicolumn{1}{c|}{$0.769 \pm 0.001$} & {$0.847 \pm 0.014$} & \multicolumn{1}{c|}{$0.594 \pm 0.035$} & {$0.900 \pm 0.034$} & \multicolumn{1}{c|}{$0.777 \pm 0.068$} & {$0.890 \pm 0.011$} & \multicolumn{1}{c|}{$0.713 \pm 0.011$} & {$0.879 \pm 0.017$} \\ 
\hline
6 & MedForge-Fusion & \multicolumn{1}{c|}{$\textbf{0.846} \pm 0.115$} & {$\textbf{0.854} \pm 0.174$} & \multicolumn{1}{c|}{$0.781 \pm 0.374$} & {$0.914 \pm 0.168$} & \multicolumn{1}{c|}{$0.792 \pm 0.084$} & {$0.908 \pm 0.114$} & \multicolumn{1}{c|}{$0.806 \pm 0.088$} & {$0.892 \pm 0.021$} \\
7 & MedForge-Mixture & \multicolumn{1}{c|}{$0.803 \pm 0.073$} & {$0.846 \pm 0.053$} & \multicolumn{1}{c|}{$\textbf{0.959} \pm 0.023$} & {$\textbf{0.998} \pm 0.001$} & \multicolumn{1}{c|}{$\textbf{0.966} \pm 0.008$} & {\textbf{0.978} $\pm$ 0.008} & \multicolumn{1}{c|}{\textbf{0.909} $\pm$ 0.023} & {\textbf{0.941} $\pm$ 0.021} \\
\bottomrule
\end{tabular}
}
\end{table*}

%% file: assets/tab/order.tex
\begin{table*}[ht]
\centering
\caption{\textbf{The Effect of Task Fusion Order.} In all three orders, the final model shows effectiveness on all the tasks, thus reflecting the robustness of our MedForge-Mixture mode. 
The order column shows the order of tasks for merging, and the capital letters (B for BreakHis, L for LC25000, M for MedFMC-Colon) refer to the tasks of breast cancer classification (BreakHis), lung cancer classification (LC25000), and colon cancer classification (MedFMC-Colon). For example, B $\rightarrow$ L $\rightarrow$ M represents the main plugin module first loads the branch plugin module trained on Breakhis, then merges with the branch plugin module for LC25000, and finally merges the MedFMC-Colon branch plugin module.}
\label{tab:order}
\resizebox{\textwidth}{!}{
\begin{tabular}{c|c|cc|cc|cc|cc}
\toprule
\multirow{2}{*}{method} & \multirow{2}{*}{order} & \multicolumn{2}{c|}{BreakHis} & \multicolumn{2}{c|}{LC25000} & \multicolumn{2}{c|}{MedFMC-Colon} & \multicolumn{2}{c}{Avg-3 tasks} \\
 & & \multicolumn{1}{c|}{ACC} & AUC & \multicolumn{1}{c|}{ACC} & AUC & \multicolumn{1}{c|}{ACC} & AUC & \multicolumn{1}{c|}{ACC} & \multicolumn{1}{c}{AUC} \\ \hline
\multirow{3}{*}{Fusion}&B$\rightarrow$L$\rightarrow$M & \multicolumn{1}{c|}{$0.641 \pm 0.141$} & {$0.685 \pm 0.100$} & \multicolumn{1}{c|}{$0.958 \pm 0.062$} & {$0.997 \pm 0.005$} & \multicolumn{1}{c|}{$0.773 \pm 0.176$} & {$0.857 \pm 0.133$} & \multicolumn{1}{c|}{$0.790 \pm 0.027$} & {$0.846 \pm 0.035$} \\
 & B$\rightarrow$M$\rightarrow$L & \multicolumn{1}{c|}{$0.846 \pm 0.115$} & {$0.854 \pm 0.174$} & \multicolumn{1}{c|}{$0.781 \pm 0.374$} & {$0.914 \pm 0.168$} & \multicolumn{1}{c|}{$0.792 \pm 0.084$} & {$0.908 \pm 0.114$} & \multicolumn{1}{c|}{$0.806 \pm 0.088$} & {$0.892 \pm 0.021$} \\
 & M$\rightarrow$L$\rightarrow$B & \multicolumn{1}{c|}{$0.680 \pm 0.064$} & {$0.699 \pm 0.155$} & \multicolumn{1}{c|}{$0.873 \pm 0.195$} & {$0.975 \pm 0.047$} & \multicolumn{1}{c|}{$0.847 \pm 0.169$} & {$0.939 \pm 0.075$} & \multicolumn{1}{c|}{$0.800 \pm 0.022$} & {$0.871 \pm 0.055$} \\
\hline
 \multirow{3}{*}{Mixture}&B$\rightarrow$L$\rightarrow$M & \multicolumn{1}{c|}{$0.698 \pm 0.044$} & {$0.755 \pm 0.058$} & \multicolumn{1}{c|}{$0.977 \pm 0.04$} & {$0.999 \pm 0.001$} & \multicolumn{1}{c|}{$0.938 \pm 0.056$} & {$0.969 \pm 0.022$} & \multicolumn{1}{c|}{$0.871 \pm 0.007$} & {$0.907 \pm 0.025$} \\
 & B$\rightarrow$M$\rightarrow$L & \multicolumn{1}{c|}{$0.803 \pm 0.073$} & {$0.846 \pm 0.053$} & \multicolumn{1}{c|}{$0.959 \pm 0.023$} & {$0.998 \pm 0.001$} & \multicolumn{1}{c|}{$0.966 \pm 0.008$} & {$0.978 \pm 0.008$} & \multicolumn{1}{c|}{$0.909 \pm 0.023$} & {$0.941 \pm 0.021$} \\
 & M$\rightarrow$L$\rightarrow$B & \multicolumn{1}{c|}{$0.791 \pm 0.098$} & {$0.823 \pm 0.094$} & \multicolumn{1}{c|}{$0.972 \pm 0.050$} & {$0.997 \pm 0.003$} & \multicolumn{1}{c|}{$0.942 \pm 0.027$} & {$0.974 \pm 0.014$} & \multicolumn{1}{c|}{$0.901 \pm 0.024$} & {$0.931 \pm 0.034$} \\
\bottomrule
\end{tabular}
}

\end{table*}

%% file: assets/tab/rebuttal.tex
\begin{table*}[ht]
\centering
\caption{\textbf{Ablation study of using different dataset distillation methods.} Compared to other dataset distillation methods, the distribution matching method shows superb performance.}
\label{tab:rebuttal}
\resizebox{\textwidth}{!}{
\begin{tabular}{c|c|cc|cc|cc|cc}
\toprule
\multirow{2}{*}{Technique} & \multirow{2}{*}{Strategy} & \multicolumn{2}{c|}{BreakHis} & \multicolumn{2}{c|}{LC25000} & \multicolumn{2}{c|}{MedFMC-Colon} & \multicolumn{2}{c}{Avg-3 tasks} \\
 &  & \multicolumn{1}{c|}{ACC} & AUC & \multicolumn{1}{c|}{ACC} & AUC & \multicolumn{1}{c|}{ACC} & AUC & \multicolumn{1}{c|}{ACC} & AUC \\ \hline
\multirow{3}{*}{DC-DSA} & Distilled data only & \multicolumn{1}{c|}{$0.795 \pm 0.064$} & {$0.898 \pm 0.023$} & \multicolumn{1}{c|}{$0.708 \pm 0.012$} & {$0.885 \pm 0.052$} & \multicolumn{1}{c|}{$0.807 \pm 0.029$} & {$0.890 \pm 0.016$} & \multicolumn{1}{c|}{$0.770 \pm 0.032$} & {$0.891 \pm 0.019$} \\
& {MedForge-Fusion} & \multicolumn{1}{c|}{$0.744 \pm 0.103$} & \textbf{$0.769 \pm 0.141$} & \multicolumn{1}{c|}{$0.983 \pm 0.027$} & {$0.999 \pm 0.001$} & \multicolumn{1}{c|}{$0.911 \pm 0.093$} & \textbf{$0.954 \pm 0.053$} & \multicolumn{1}{c|}{$0.879 \pm 0.039$} & {$0.907 \pm 0.058$} \\
 & {MedForge-Mixture} & \multicolumn{1}{c|}{$0.859 \pm 0.064$} & \textbf{$0.873 \pm 0.095$} & \multicolumn{1}{c|}{$0.798 \pm 0.333$} & {$0.904 \pm 0.190$} & \multicolumn{1}{c|}{$0.967 \pm 0.006$} & {$0.979 \pm 0.008$} & \multicolumn{1}{c|}{$0.875 \pm 0.088$} & {$0.919 \pm 0.051$} \\ \hline
\multirow{3}{*}{DM-DSA (ours)} & Distilled data perf & \multicolumn{1}{c|}{$0.881 \pm 0.006$} & {$0.883 \pm 0.039$} & \multicolumn{1}{c|}{$0.911 \pm 0.002$} & {$0.974 \pm 0.001$} & \multicolumn{1}{c|}{$0.799 \pm 0.011$} & {$0.872 \pm 0.016$} & \multicolumn{1}{c|}{$0.864 \pm 0.002$} & {$0.910 \pm 0.010$} \\
& MedForge-Fusion & \multicolumn{1}{c|}{$\textbf{0.846} \pm 0.115$} & {$\textbf{0.854} \pm 0.174$} & \multicolumn{1}{c|}{$0.781 \pm 0.374$} & {$0.914 \pm 0.168$} & \multicolumn{1}{c|}{$0.792 \pm 0.084$} & {$0.908 \pm 0.114$} & \multicolumn{1}{c|}{$0.806 \pm 0.088$} & {$0.892 \pm 0.021$} \\
 & MedForge-Mixture & \multicolumn{1}{c|}{$0.803 \pm 0.073$} & {$0.846 \pm 0.053$} & \multicolumn{1}{c|}{$\textbf{0.959} \pm 0.023$} & {$\textbf{0.998} \pm 0.001$} & \multicolumn{1}{c|}{$\textbf{0.966} \pm 0.008$} & {\textbf{0.978} $\pm$ 0.008} & \multicolumn{1}{c|}{\textbf{0.909} $\pm$ 0.023} & {\textbf{0.941} $\pm$ 0.021} \\ \hline
\multirow{3}{*}{DM} & Distilled data perf & \multicolumn{1}{c|}{$0.859 \pm 0.026$} & {$0.889 \pm 0.021$} & \multicolumn{1}{c|}{$0.891 \pm 0.02$} & {$0.967 \pm 0.007$} & \multicolumn{1}{c|}{$0.797 \pm 0.009$} & {$0.872 \pm 0.007$} & \multicolumn{1}{c|}{$0.849 \pm 0.004$} & {$0.910 \pm 0.007$} \\
& {MedForge-Fusion} & \multicolumn{1}{c|}{$0.782 \pm 0.141$} & {$0.782 \pm 0.162$}& \multicolumn{1}{c|}{$0.950 \pm 0.017$} & {$0.994 \pm 0.005$} & \multicolumn{1}{c|}{$0.690 \pm 0.133$} & {$0.797 \pm 0.276$} & \multicolumn{1}{c|}{\textbf{$0.808 \pm 0.030$}} & \textbf{$0.858 \pm 0.038$} \\
 & {MedForge-Mixture} & \multicolumn{1}{c|}{$0.833 \pm 0.064$} & {$0.856 \pm 0.072$}& \multicolumn{1}{c|}{$0.980 \pm 0.026$} & {$0.991 \pm 0.011$} & \multicolumn{1}{c|}{$0.701 \pm 0.062$} & {$0.830 \pm 0.086$} & \multicolumn{1}{c|}{$0.838 \pm 0.006$} & {$0.892 \pm 0.033$} \\
\bottomrule
\end{tabular}}
\end{table*}

%% file: assets/tab/lora.tex
\begin{table*}[ht]
\centering
\caption{\textbf{The results of replacing LoRA with DoRA.} After replacing plugin module from LoRA to DoRA, MedForge still outperforms.}
\label{tab:dora}
\resizebox{\textwidth}{!}{
\begin{tabular}{c|cc|cc|cc|cc}
\toprule
\multirow{2}{*}{Method} & \multicolumn{2}{c|}{BreakHis} & \multicolumn{2}{c|}{LC25000} & \multicolumn{2}{c|}{MedFMC-Colon} & \multicolumn{2}{c}{Avg-3 tasks} \\
 & \multicolumn{1}{c|}{ACC} & AUC & \multicolumn{1}{c|}{ACC} & AUC & \multicolumn{1}{c|}{ACC} & AUC & \multicolumn{1}{c|}{ACC} & AUC \\ \hline
DoRA supervised & \multicolumn{1}{c|}{$0.921 \pm 0.004$} & {$0.926 \pm 0.019$} & \multicolumn{1}{c|}{$0.996 \pm 0.001$} & {$0.999 \pm 0.001$} & \multicolumn{1}{c|}{$0.970 \pm 0.001$} & {$0.995 \pm 0.000$} & \multicolumn{1}{c|}{$0.962 \pm 0.001$} & {$0.974 \pm 0.006$} \\
Distilled data only & \multicolumn{1}{c|}{$0.870 \pm 0.047$} & {$0.894 \pm 0.009$} & \multicolumn{1}{c|}{$0.903 \pm 0.006$} & {$0.972 \pm 0.002$} & \multicolumn{1}{c|}{$0.803 \pm 0.012$} & {$0.872 \pm 0.013$} & \multicolumn{1}{c|}{$0.858 \pm 0.013$} & {$0.913 \pm 0.007$} \\
\hline
LoRAHub w/ distill & \multicolumn{1}{c|}{$0.769 \pm 0.077$} & {$\textbf{0.859} \pm 0.044$} & \multicolumn{1}{c|}{$0.778 \pm 0.051$} & {$0.944 \pm 0.043$} & \multicolumn{1}{c|}{$0.798 \pm 0.026$} & {$0.880 \pm 0.011$} & \multicolumn{1}{c|}{$0.781 \pm 0.018$} & {$0.894 \pm 0.011$} \\
LoRAHub w/o distill & \multicolumn{1}{c|}{$0.744 \pm 0.026$} & {$0.845 \pm 0.023$} & \multicolumn{1}{c|}{$0.884 \pm 0.069$} & {$0.982 \pm 0.013$} & \multicolumn{1}{c|}{$0.718 \pm 0.060$} & {$0.783 \pm 0.080$} & \multicolumn{1}{c|}{$0.782 \pm 0.006$} & {$0.870 \pm 0.029$} \\
ModelSoup & \multicolumn{1}{c|}{$0.731 \pm 0.039$} & {$0.850 \pm 0.016$} & \multicolumn{1}{c|}{$0.626 \pm 0.040$} & {$0.906 \pm 0.036$} & \multicolumn{1}{c|}{$0.802 \pm 0.037$} & {$0.909 \pm 0.003$} & \multicolumn{1}{c|}{$0.720 \pm 0.018$} & {$0.888 \pm 0.018$} \\
\hline
MedForge-Fusion & \multicolumn{1}{c|}{$0.736 \pm 0.033$} & {$0.849 \pm 0.008$} & \multicolumn{1}{c|}{$0.828 \pm 0.056$} & {$0.967 \pm 0.008$} & \multicolumn{1}{c|}{$0.792 \pm 0.049$} & {$0.867 \pm 0.037$} & \multicolumn{1}{c|}{$0.785 \pm 0.018$} & {$0.894 \pm 0.013$} \\
MedForge-Mixture & \multicolumn{1}{c|}{$\textbf{0.830} \pm 0.099$} & {$0.842 \pm 0.043$} & \multicolumn{1}{c|}{$\textbf{0.926} \pm 0.071$} & {$\textbf{0.996} \pm 0.003$} & \multicolumn{1}{c|}{$\textbf{0.962} \pm 0.010$} & {$\textbf{0.974} \pm 0.003$} & \multicolumn{1}{c|}{$\textbf{0.906} \pm 0.016$} & {$\textbf{0.937} \pm 0.014$} \\
\bottomrule
\end{tabular}}
\end{table*}

%% file: assets/img/feature.tex
\begin{figure*}[t]
    \centering
    \includegraphics[width=\textwidth]{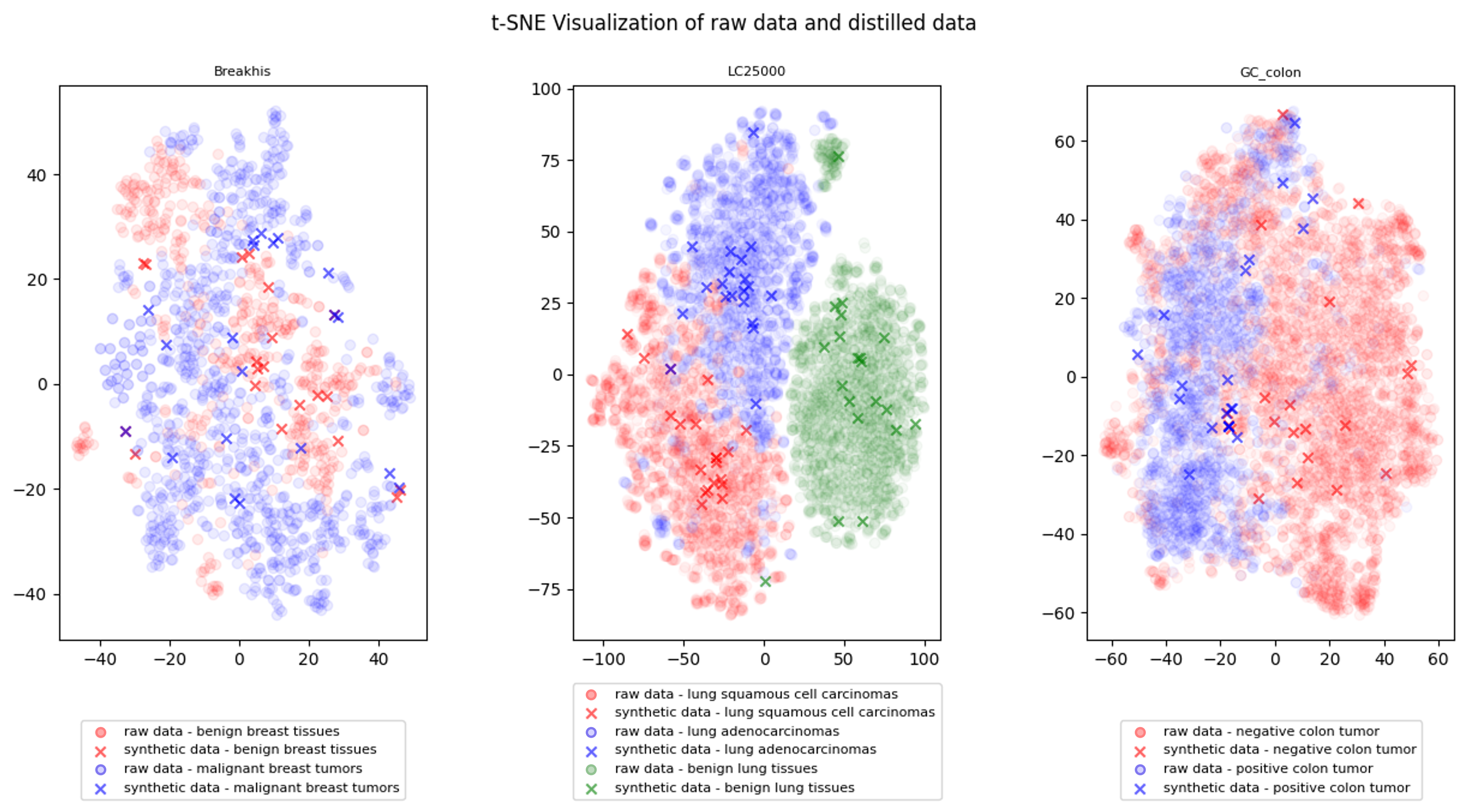}
    \caption{\textbf{The distribution of raw dataset and distilled data.}}
    \label{fig:dist_all}
\end{figure*}

%% file: sections/6-ablation.tex
\section{Ablation Study}
\subsection{The Effect of Task Merging Order}
In real-world applications, contributors can push their branch modules and distilled data anytime, leading to various fusion orders. The order of merging operations might introduce randomness and affect the robustness of overall performance. Thus, it is necessary to determine the effect of task merging order on model performance. In this ablation study, we shuffle the order of tasks and evaluate the model performance with multiple order combinations.

As shown in Table~\ref{tab:order}, MedForge could consistently achieve good performance on all three tasks (e.g., for MedForge-Mixture, B $\rightarrow$ M $\rightarrow$ L versus M $\rightarrow$ L $\rightarrow$ B, 0.941 versus 0.931 in AUC). The different orders could cause various model performances due to the inherent characteristics of the datasets. For example, BreakHis has the smallest training set, leading to greater sensitivity to the merging order. Moreover, observing the results across three tasks and the averaged performance, MedForge-Mixture consistently demonstrates greater robustness compared to the MedForge-Fusion approach. This superior performance is achieved by maintaining the integrity of each plugin module’s internal parameters and summing their outputs rather than altering the parameters themselves. This approach ultimately enhances the model's robustness and stability.

\subsection{The Effect of Dataset Distillation} We assessed model performance using distilled data achieved by different dataset distillation strategies. We have extended to Dataset Condensation (DC) approaches and explored the significance of the Differentiable Siamese Augmentation (DSA) technique. From Table~\ref{tab:rebuttal}, we demonstrate that our distillation module (DM-DSA) achieves strong performances among datasets in average (ACC: \textbf{0.909 (DM-DSA, MixMerge)} vs. 0.875 (DC-DSA) and 0.838 (DM), AUC: \textbf{0.941 (DM-DSA, ours)} vs. 0.919 (DC-DSA) and 0.892 (DM)). Moreover, the average ACC of Mixture exceeds all baselines, indicating the superiority of MedForge. The results reveal the characteristics of different distillation methods, as the distribution matching (DM) method inclines to learn the similar distribution of raw dataset, thus becoming crucial guidance in coefficients learning at the merging stage. Meanwhile, the Differentiable Siamese Augmentation (DSA) technique further enhances the representativeness of distilled data by providing more efficient and robust feature learning.

\subsection{The Effect of Different Plugin Module}
To analyze the flexibility of MedForge by using different plugin module adoptions, we evaluate MedForge with recently proposed DoRA~\cite{liu2024dora} (weight-decomposed low-rank adaptation) in our main experiments. DoRA is strong at decomposing pre-trained weights into magnitude and direction components for model fine-tuning. DoRA improves learning capacity and training stability while minimizing trainable parameters and inference costs. 
From Table~\ref{tab:dora}, we demonstrated that MedForge can yield a consistently strong performance when using different PEFT methods (e.g., DoRA as the plugin module). While using DoRA as a plugin module, MedForge still achieves the leading performance (Mixture: ACC: 0.906, AUC: 0.937) compared to other merging strategies. Meanwhile, medForge-Mixture shows positive performance that highlights the advantage of mixture approach in maintaining the integrity of parameters within each plugin module towards enhancing model fusion performance. Overall, the effectiveness of MedForge in replacing DoRA as a plugin module demonstrates the potential of MedForge approach across different experimental settings.

\subsection{Distilled Dataset Quality Verification} We employ t-SNE to visualize the high-dimensional feature embedding of the distilled dataset and the original dataset into a two-dimensional space. Such visualization is helpful for assessing the representativeness of the distilled data concerning the original data distribution. The visualized t-SNE plots revealed that, despite using only a small number of sample points, the distilled dataset features closely approximated the distribution of the original data, highlighting the distilled data's high representativeness and informative nature. From (\ref{fig:dist_all}), we assessed the degree of fidelity with which the distilled dataset captured the essential characteristics of the original data. This design provides insights into the effectiveness of our data distillation process and the validity of using the distilled dataset to guide the optimization of the model coefficients.

%% file: sections/7-discussion.tex
\section{Discussion}
The development of foundation models has increasingly relied on accessible data support to address complex tasks~\cite{zhang2024data}. Yet major challenges remain in collecting scalable clinical data in the healthcare system, such as data silos and privacy concerns. To overcome these challenges, MedForge integrates multi-center clinical knowledge sources into a cohesive medical foundation model via a collaborative scheme. MedForge offers a collaborative path to asynchronously integrate multi-center knowledge while maintaining strong flexibility for individual contributors.
This key design allows a cost-effective collaboration among clinical centers to build comprehensive medical models, enhancing private resource utilization across healthcare systems.

Inspired by collaborative open-source software development~\cite{raffel2023building, github}, our study allows individual clinical institutions to independently develop branch modules with their data locally. These branch modules are asynchronously integrated into a comprehensive model without the need to share original data, avoiding potential patient raw data leakage. Conceptually similar to the open-source collaborative system, iterative module merging development ensures the aggregation of model knowledge over time while incorporating diverse data insights from distributed institutions. In particular, this asynchronous scheme alleviates the demand for all users to synchronize module updates as required by conventional methods (e.g., LoRAHub~\cite{huang2023lorahub}).

MedForge's framework addresses multiple data challenges in the cycle of medical foundation model development, including data storage, transmission, and leakage. As the data collection process requires a large amount of distributed data, we show that dataset distillation contributes greatly to reducing data storage capacity. In MedForge, individual contributors can simply upload a lightweight version of the dataset to the central model developer. As a result, the remarkable reduction in data volume (e.g., 175 times less in LC25000) alleviates the burden of data transfer among multiple medical centers. For example, we distilled a 10,500 image training set into 60 representative distilled data while maintaining a strong model performance. We choose to use a lightweight dataset as a transformed representation of raw data to avoid the leakage of sensitive raw information.
Second, the asynchronous collaboration mode in MedForge allows flexible model merging, particularly for users from various local medical centers to participate in model knowledge integration. 
Third, MedForge reformulates the conventional top-down workflow of building foundational models by adopting a bottom-up approach. Instead of solely relying on upstream builders to predefine model functionalities, MedForge allows medical centers to actively contribute to model knowledge integration by providing plugin modules (i.e., LoRA) and distilled datasets. This approach supports flexible knowledge integration and allows models to be applicable to wide-ranging clinical tasks, addressing the key limitation of fixed functionalities in traditional workflows.

We demonstrate the strong capacity of MedForge via the asynchronous merging of three image classification tasks. MedForge offered an incremental merging strategy that is highly flexible compared to plain parameter average~\cite{wortsman2022model} and LoRAHub~\cite{huang2023lorahub}. Specifically, plain parameter averaging merges module parameters directly and ignores the contribution differences of each module. Although LoRAHub allows for flexible distribution of coefficients among modules, it lacks the ability to continuously update, limiting its capacity to incorporate new knowledge during the merging process. In contrast, MedForge shows its strong flexibility of continuous updates while considering the contribution differences among center modules. The robustness of MedForge has been demonstrated by shuffling merging order (Tab~\ref{tab:order}), which shows that merging new-coming modules will not hurt the model ability of previous tasks in various orders, mitigating the model catastrophic forgetting. 
MedForge also reveals a strong generality on various choices of component modules. Our experiments on dataset distillation settings (such as DC and without DSA technique) and PEFT techniques (such as DoRA) emphasize the extensible ability of MedForge's module settings. 

To fully exploit multi-scale clinical data, it will be necessary to include broader data modalities (e.g., electronic health records and radiological images). Managing these diverse data formats and standards among numerous contributors can be challenging due to the potential conflict between collaborators. 
Moreover, since MedForge integrates multiple clinical tasks that involve varying numbers of classification categories, conventional classifier heads with fixed class sizes are not applicable. However, the projection head of the CLIP model, designed to calculate similarities between image and text, is well-suited for this scenario. It allows MedForge to flexibly handle medical datasets with different category numbers, thus overcoming the challenge of multi-task classification. That said, this design choice also limits the variety of model architectures that can be utilized, as it depends specifically on the CLIP framework. Future investigations will explore extensive solutions to make the overall architecture more flexible. Additionally, incorporating more sophisticated data anonymization, such as synthetic data generation~\cite{ding2023large}, and encryption methods can also be considerable. To improve data privacy protection, test-time adaptation technique~\cite{wang2020tent, liang2024comprehensive} without substantial training data can be considered to alleviate the burden of data sharing in the healthcare system.

%% file: sections/8-conclusion.tex
\section{Conclusion}
Medical community faces long-standing challenges in integrating decentralized, privacy-se
nsitive data at scale. In this study, we have proposed MedForge as a community-driven approach for medical foundation model development. MedForge enables asynchronous collaboration among multiple institutions to aggregate medical knowledge on various tasks. MegForge allows continuous model updates from separate contributors while effectively protecting the in-house patient data privacy. MedForge reveals strong performance across multiple tasks, including breast cancer, lung cancer subtypes, and colon cancer classification, showing the robustness of merging order and model architecture. These major findings demonstrate the efficiency of MedForge in wide-ranging medical applications toward multi-center performance evaluation.